\pdfoutput=1

\documentclass[11pt]{article}

\usepackage{ACL2023}
\defcitealias{rajpoot-parikh-2023-gpt}{Rajpoot et al.}
\defcitealias{ettaleb-etal-2025-contribution}{Ettaleb et al.}
\defcitealias{Borchert2023COREAF}{Borchert et al.}
\defcitealias{aguda-etal-2024-large}{Aguda et al.}
\defcitealias{refindpaper}{Kaur et al.}
\defcitealias{Efeoglu2024RelationEW}{Efeoglu et al.}
\defcitealias{Guo2025BridgingGA}{Guo et al.}
\defcitealias{Wu2019EnrichingPL}{Wu et al.}

\usepackage{hyperref}
\usepackage{times}
\usepackage{latexsym}
\usepackage{graphicx}
\usepackage{booktabs}
\usepackage[T1]{fontenc}

\usepackage[utf8]{inputenc}

\usepackage{microtype}

\usepackage{inconsolata}

\title{Comparative Analysis of AI Agent Architectures for Entity Relationship Classification}

\author{Maryam Berijanian \\
  Michigan State University \\
  \texttt{berijani@msu.edu} \\\And
  Kuldeep Singh \\
  Michigan State University \\
  \texttt{singhku2@msu.edu} \\\And 
  Amin Sehati \\
  \texttt{amin@aminsehati.com} 
  }

\begin{document}
\maketitle
\begin{abstract}
Entity relationship classification remains a challenging task in information extraction, especially in scenarios with limited labeled data and complex relational structures. In this study, we conduct a comparative analysis of three distinct AI agent architectures designed to perform relation classification using large language models (LLMs). The agentic architectures explored include (1) reflective self-evaluation, (2) hierarchical task decomposition, and (3) a novel multi-agent dynamic example generation mechanism, each leveraging different modes of reasoning and prompt adaptation. In particular, our dynamic example generation approach introduces real-time cooperative and adversarial prompting. We systematically compare their performance across multiple domains and model backends. Our experiments demonstrate that multi-agent coordination consistently outperforms standard few-shot prompting and approaches the performance of fine-tuned models. These findings offer practical guidance for the design of modular, generalizable LLM-based systems for structured relation extraction. The source codes and dataset are available at \href{https://github.com/maryambrj/ALIEN.git}{https://github.com/maryambrj/ALIEN.git}.
\end{abstract}

\section{Introduction}
Large language models (LLMs) have shown exceptional capabilities in general-purpose language understanding, reasoning, and generation. However, effectively harnessing these capabilities for relation extraction—particularly in specialized domains like scientific literature or financial documents—remains a significant challenge \cite{Song2025InjectingDK}. Tasks involving nuanced relational semantics often demand more structured, interpretable, and context-aware decision-making pipelines \cite{Dunn2024StructuredIE}.



In this work, we explore and compare three agent-based architectures for relation classification, each incorporating distinct reasoning strategies to enhance performance in n-shot settings. These include hierarchical task delegation, iterative self-critique, and dynamic example generation; mechanisms that coordinate large language model (LLM) behavior through structured workflows. Rather than proposing a unified framework, we evaluate these strategies individually across diverse datasets to better understand their impact on classification accuracy, generalization, and interpretability. Our analysis emphasizes how different forms of agentic coordination affect model performance, especially in low-resource and zero-shot scenarios.



To better understand how different agentic strategies influence relational understanding in LLMs, we conduct a comprehensive empirical study comparing three distinct multi-agent architectures. Our work goes beyond introducing a new system—it provides actionable insights into how design choices in agent-based prompting affect relation classification performance across domains.
Our key contributions are as follows:

\begin{enumerate}


\item Systematic Comparison of Agent Architectures: We conduct a structured evaluation of three modular agent-based architectures—\textit{Hierarchical Multi-Agent}, \textit{Generator-Reflection}, and \textit{Dynamic-Example Agent}—each embodying distinct strategies for task supervision, delegation, and critique. While these mechanisms have been explored individually in prior work, our study is the first to comparatively assess their effectiveness for the relation classification task using large language models.

\item Multi-Domain Evaluation with State-of-the-Art Models: We evaluate all three architectures using Gemini 2.5 Flash and GPT-4o across three datasets representing different domains: financial (REFinD) \cite{refindpaper}, scientific (CORE) \cite{Borchert2023COREAF}, and general-domain (SemEval 2010 Task 8) \cite{Hendrickx2010SemEval2010T8}. Our results highlight consistent performance gains over standard few-shot prompting and show the potential of agent-based coordination to approach or match fine-tuned model performance.
\end{enumerate}



\paragraph{Related Work}
Conventional relation extraction methods rely heavily on supervised learning with closed-label sets \cite{zeng2014relation, miwa2016end, zhou2016attention}. Transformer-based models such as BERT \cite{devlin2019bert} and RoBERTa \cite{liu2019roberta} have improved performance, especially when augmented with distant supervision \cite{riedel2010modeling}, but struggle to adapt flexibly to new relation types or domains. Recently, instruction tuning and in-context learning of LLMs like GPT-4 \cite{openai2023gpt4}, LLaMA-3 \cite{Dubey2024TheL3}, and Gemini \cite{team2023gemini} has become a popular alternative \cite{brown2020language, sanh2022multitask}, though these approaches still lack mechanisms for targeted specialization and interactive correction.


Multi-agent frameworks have emerged as promising tools for decomposing tasks across specialized agents, enabling collaborative reasoning \cite{shinn2023reflexion, li2024camel, mialon2023augmented}. However, few efforts have adapted multi-agent methods explicitly to relation extraction, and most lack real-time interaction between agents for dynamic supervision or example construction. For example, prior work has explored document-level relation extraction using graph reasoning \cite{hou2024multiagent} or modular retrieval-memory-extraction \cite{Shi2024AgentREAA}, but these systems employ static prompts or manually selected examples.

Recent research has investigated using LLMs to generate diverse training examples for data augmentation. For instance, Li et al. (2025) focus on improving sample diversity via prompt instructions and preference optimization, while Liu et al. (2024) employ a self-prompting approach for zero-shot relation extraction. However, these methods rely on single-agent, static generation without real-time adaptation. In contrast, we introduce a novel multi-agent approach for dynamic example generation, where cooperative and adversarial agents construct positive and negative samples tailored to each test instance. To our knowledge, this is the first application of agent-mediated, per-instance dynamic example generation in relation extraction.



\section{Methods}
In this section, we describe the datasets, baseline prompting strategies, and three agent-based architectures developed for relation classification.

\subsection{Datasets}
\label{datasets}

We evaluate our architectures on three datasets, chosen for their diversity in domain, entity types, and relation complexity:

REFinD \cite{refindpaper} is a financial relation classification dataset. It includes 19 unique relation types across entity pairs like companies, dates, and financial metrics. The language is formal and domain-specific, with longer sentence lengths (avg. 53.7 tokens) than general-domain datasets. The dataset contains roughly 29,000 annotated instances, with standard splits for training, validation, and testing.

CORE \cite{Borchert2023COREAF} is a scientific corpus extracted from research papers. It focuses on relationships between authors, affiliations, and research topics. The dataset includes academic-specific relations such as \texttt{affiliated\_with}, \texttt{published\_in}, and \texttt{works\_on}. Sentences are rich in technical language and abbreviations, making classification more challenging.

SemEval 2010 Task 8 \cite{Hendrickx2010SemEval2010T8} is a widely used benchmark dataset for relation classification. It covers nine directed semantic relations between nominals such as \texttt{Cause-Effect}, \texttt{Component-Whole}, and \texttt{Instrument-Agency}. The language is general-domain and the dataset is balanced, enabling comparative evaluation across systems.

Each dataset presents distinct challenges in terms of vocabulary, syntax, and domain specificity, providing a robust testbed for evaluating the adaptability and generalization capability of our agent-based architectures.

\subsection{Few-Shot Prompting Baselines}

To establish baseline performance and enable meaningful comparisons with our agent-based approaches, we conducted few-shot prompting experiments using GPT-4o \cite{Hurst2024GPT4oSC} and Gemini 2.5 Flash models \cite{gemini25} across all three datasets. We implemented zero-shot, 3-shot, and 5-shot learning scenarios. No iterative feedback, reflection, or agent-based routing was employed in these baselines. 

\subsection{Generator-Reflection Architecture}

\begin{figure}[t]
    \centering
    \includegraphics[width=0.255\textwidth]{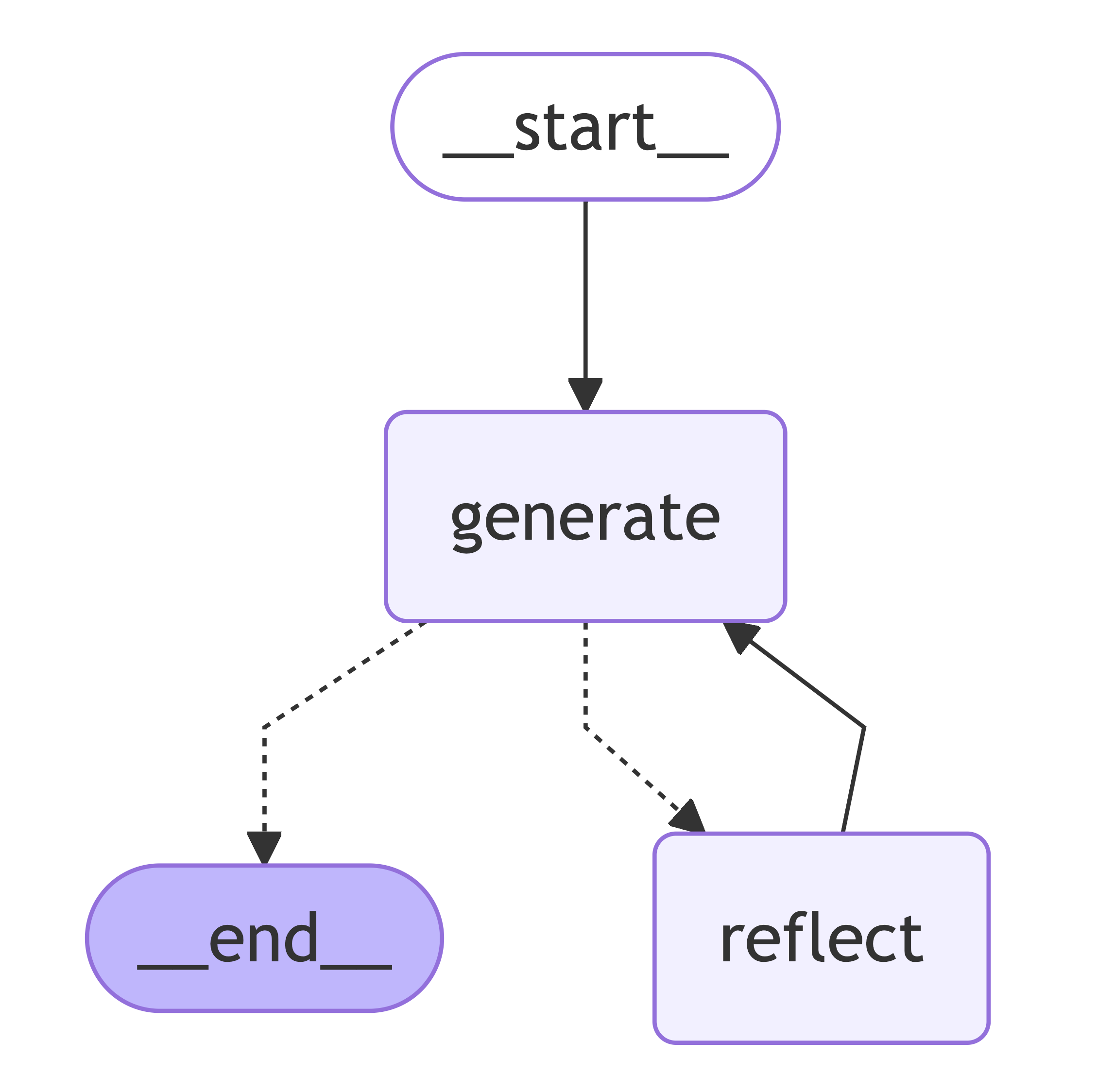}
    \caption{The Generator-Reflection architecture.}
    \label{fig:reflection_architecture}
\end{figure}

We used a Generator-Reflection architecture, comprising two cooperative agents: the Generator Agent, responsible for initial relation classification, and the Reflection Agent, for critiquing and refining these classifications iteratively (Figure \ref{fig:reflection_architecture}).


\subsubsection{Generator Agent}

The Generator Agent is tasked with performing the initial classification of relationships between entity pairs, leveraging context provided at the sentence level. The agent receives structured input and the associated contextual sentence describing their interaction. The Generator Agent outputs relationship labels for each entity pair, strictly selected from a predefined set of permissible relation types.


\subsubsection{Reflection Agent}

The Reflection Agent functions as a quality assurance mechanism, critically evaluating the Generator Agent's outputs. This agent meticulously assesses the generated tables and provides critiques and suggestions for corrections, thereby guiding the Generator Agent toward improved accuracy in subsequent iterations.


\subsubsection{Feedback Loop \& Termination Criteria}

The Generator and Reflection agents operate within an iterative feedback loop. Each iteration comprises three primary steps:

\begin{enumerate}
\item Generation Phase: The Generator Agent creates an initial or revised relation extraction table.
\item Reflection Phase: The Reflection Agent critically reviews the table, offering specific feedback.
\item Revision Decision: The system determines whether further refinement is necessary, based on whether actionable critiques are identified.
\end{enumerate}

The iterative loop is terminated when a maximum threshold of three dialogue cycles (iterations), in this case three times, has been reached to prevent indefinite processing.







\subsection{Hierarchical Multi-Agent Architecture}

\begin{figure}[t]
    \centering
    \includegraphics[width=0.45\textwidth]{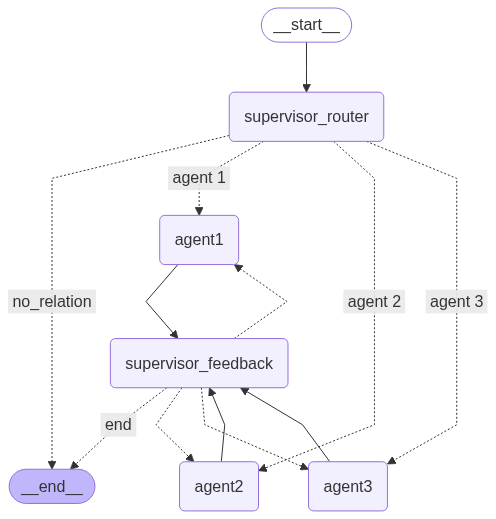}
    \caption{The Hierarchical Multi-Agent architecture.}
    \label{fig:supervisor_architecture}
\end{figure}

The architecture consists of one Orchestrator agent responsible for routing classification tasks, and multiple Specialist agents, each specialized in distinct relation categories (Figure~\ref{fig:supervisor_architecture}).

\subsubsection{Orchestrator Routing Mechanism}

The Orchestrator Agent is designed to analyze input sentences and dynamically route them to the most appropriate Specialist agent. This decision-making process is guided by structured prompts containing explicit descriptions of agent specializations and the associated relation labels each agent can handle. Each dataset was associated with unique mappings of agent specializations and permissible labels. If no relevant relation was evident, the Orchestrator explicitly assigned the sentence to the \texttt{no\_relation} category, thereby bypassing further agent involvement.


\subsubsection{Specialist Agents}

Three Specialist agents were employed within each dataset-specific run, with each agent specializing in a distinct domain of relationship classification. Agent specializations and their permissible relation labels varied among datasets. For example, for the CORE dataset, agents were specialized as follows:

\begin{itemize}
\item Agent 1 (Corporate Structure): Specialized in ownership and corporate structural relations (e.g., \texttt{acquired\_by}, \texttt{subsidiary\_of}, \texttt{shareholder\_of}).
\item Agent 2 (Business Interactions): Specialized in peer-to-peer business interactions (e.g., \texttt{competitor\_of}, \texttt{collaboration}, \texttt{client\_of}).
\item Agent 3 (Market and Regulatory Context): Specialized in product and market-context relations (e.g., \texttt{product\_or\_service\_of}, \texttt{regulated\_by}, \texttt{traded\_on}).
\end{itemize}


Each Specialist agent received structured prompts which constrained agent responses strictly to their predefined label sets.


\subsubsection{Orchestrator Feedback Loop}

After a Specialist agent provides an initial classification, the Orchestrator evaluates the correctness and appropriateness of the predicted label. This evaluation involves an explicit verification step where the Orchestrator assesses whether the Specialist agent’s proposed relationship aligns with its pre-defined specialization and permissible relation labels. If the Orchestrator detects a misclassification or label inconsistency, it issues explicit feedback instructing the Specialist agent to reconsider and select an alternative relation label. This iterative interaction continues under clearly defined termination conditions:

\begin{itemize}
\item When the Orchestrator confirms the accuracy of the Specialist agent’s classification.
\item If the maximum limit of three classification attempts per instance is reached.
\item If the Specialist agent repeatedly returns the \texttt{no\_relation} classification, indicating the absence of a meaningful relation after two attempts.
\end{itemize}






\subsection{Dynamic-Example Generator Agent}
\begin{figure*}[t]
    \centering
    \includegraphics[width=0.7\textwidth]{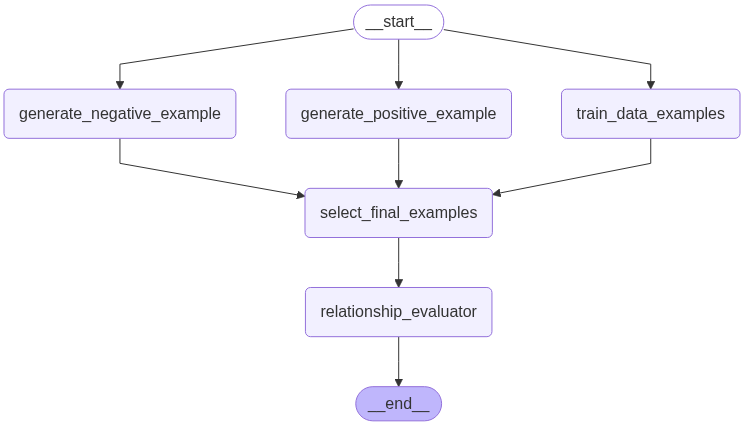}
    \caption{The Dynamic-Example Generator Agent.}
    \label{fig:dyn-ex}
\end{figure*}

We developed a Dynamic-Example Generator Agent (Figure \ref{fig:dyn-ex}) that operates in three steps: it first constructs a diverse pool of examples, then selects a subset to form a tailored context, and finally performs relation classification—enhancing adaptability, robustness, and accuracy.

\subsubsection{Example Pool Creation}
To build high-quality, context-specific prompts, we combine examples from multiple sources. For each input, the Dynamic Example Generator Agent produces positive and adversarial negative examples to capture both prototypical and challenging cases. Simultaneously, semantically similar examples are retrieved from training data using a retrieval index. This curated mix of reinforcing, contrastive, and semantically similar examples provides a wide variety of contextual signals that can be effectively leveraged to improve classification accuracy and generalizability.

\subsubsection{Dynamic Example Selection}
The example selection step evaluates generated examples against the input to choose the most relevant ones. This ensures the model receives a focused, informative context that enhances classification. Selection criteria include relevance, diversity of relation types, and the potential to support or challenge the model’s understanding.

\subsubsection{Classification Process}
In the final classification step, the model processes a dynamically assembled prompt—comprising the selected examples and input instance—to predict the relation label, leveraging enriched context from both generated and training data examples.


\subsection{Experiment Setup}
We conducted extensive experiments across the three datasets evaluating each of our agent-based architectures under multiple configurations. These configurations varied in terms of model pairings and prompting strategies, including N-shot prompting. The goal was to assess the effectiveness of hierarchical supervision, iterative feedback, and dynamic example generation in enhancing relation classification accuracy.

\subsubsection{Prompting and Model Setup}
For few-shot prompting, we used randomly selected (but fixed) examples from the training data in 3-shot and 5-shot configurations. Prompts included structured inputs comprising entity pairs and contextual sentences designed to support accurate classification.

For the Hierarchical Multi-Agent architecture, we conducted comparative experiments using two model configurations, with the orchestrator allowed a maximum of three classification attempts per input instance: (1) Gemini-2.5 Flash as the orchestrator with GPT-4o as the specialist agents, and (2) GPT-4o-mini as the orchestrator with GPT-4.1-mini as the specialists.

For the Generator-Reflection Architecture, we limited the iterative refinement to a maximum of three dialogue cycles to ensure computational efficiency. Two model setups were evaluated across all datasets: (1) Gemini-2.5 Flash as the Generator and GPT-4o as the Reflector, and (2) Gemini-2.5 Pro as the Generator and GPT-4o-mini as the Reflector.

For Dynamic-Example Generator Agent, we generated 10 positive and 10 negative examples per input instance, along with sampling 20 examples from the training set. From this pool, 10 examples were selected for inclusion in the final prompt. To retrieve relevant training examples, we used FAISS with embeddings generated by OpenAI’s text-embedding-3-small model, enabling efficient semantic similarity search. Gemini 2.5 Flash was used for both generating the candidate examples and performing the final relation classification, while GPT-4o was responsible for selecting the most contextually appropriate examples from the candidate pool.

All configurations were implemented using LangGraph and tested across the REFinD, CORE, and SemEval 2010 Task 8 datasets.

\subsubsection{Evaluation Metrics}
Classification performance was evaluated using standard metrics, including macro-average F1 and micro-average F1 scores, ensuring a comprehensive assessment of the model's predictive capabilities across diverse relation classes.

\section{Results and Discussion}
Table~\ref{tab:comparison_prior_work1} presents relation classification performance on the CORE dataset. Our Generator-Reflection architecture achieves an F1 Micro of 0.76, demonstrating strong performance that is competitive with or exceeds several prior methods. Notably, our approach surpasses prompting-based baselines such as Llama-3 and BERT-Prompt. Also in the case of BERT-Prompt, the evaluation was on only a subset of the test set. While fine-tuned models report a slightly higher F1 Micro of 0.80, they benefit from additional task-specific training. This highlights the effectiveness and generality of our multi-agent architecture, which achieves high performance without relying on fine-tuning or extensive supervision.

\begin{table}[t]
\centering
\small
\caption{Relation classification performance on the CORE dataset. Bold rows are our results. \\
G-2.5-P = Gemini 2.5 Pro; Gen-Ref = Generator-Reflection architecture.}
\label{tab:comparison_prior_work1}
\begin{tabular}{llcc}
\toprule
\textbf{Model} & \textbf{Setting} & \textbf{F1-Mac} & \textbf{F1-Mic} \\
\midrule
Llama-3 & 0-Shot & -- & 0.69 \\
\citepalias{ettaleb-etal-2025-contribution} & 3-Shot & -- & 0.70 \\
 & Fine-tuning & -- & 0.80 \\
 \midrule
BERT-Prompt$^1$ & 5-Way 1-Shot$^1$&  0.67& 0.53 \\
\citepalias{Borchert2023COREAF} &  5-Way 5-Shot$^1$& 0.88 & 0.74 \\
\midrule
\textbf{G-2.5-P,GPT-o3} &\textbf{Gen-Ref} & \textbf{0.77} & \textbf{0.76} \\
\textbf{G-2.5-F,GPT-4o} & \textbf{Hier-Multi} & \textbf{0.74} & \textbf{0.73} \\
\textbf{G-2.5-F,GPT-4o} & \textbf{Dyn-Ex} & \textbf{0.75} &  \textbf{0.72}\\
\bottomrule
\end{tabular}
\vspace{1mm}
\raggedright
\footnotesize{1) Evaluated on a subset of the full test set, not the complete CORE benchmark.}
\end{table}

Table~\ref{tab:comparison_prior_work2} presents relation classification results on the REFinD dataset. Our Generator-Reflection architecture achieves an F1 Micro of 0.61, showing strong performance despite not leveraging task-specific pretraining or optimized few-shot prompts. While PaLM 2 and GPT-4 report higher scores in few-shot settings, their setups include carefully curated in-context examples or use significantly more examples, which limits the fairness of a direct comparison. Similarly, Luke-base achieves the highest F1 Micro of 0.75 but benefits from pretraining explicitly for relation extraction. In contrast, our approach relies on general-purpose LLMs and simple prompt configurations, yet delivers competitive results, underscoring the versatility and robustness of our multi-agent design.

\begin{table}[t]
\centering
\small
\caption{Relation classification performance on the REFinD dataset. Bold rows are our results. \\
G-2.5-F = Gemini 2.5 Flash; Gen-Ref = Generator-Reflection architecture.}
\label{tab:comparison_prior_work2}
\begin{tabular}{llcc}
\toprule
\textbf{Model} & \textbf{Setting} & \textbf{F1-Mac} & \textbf{F1-Mic} \\
\midrule
Luke-base$^1$ & Pretrained  & 0.56 & 0.75 \\
\citepalias{refindpaper} & for RE$^1$ &  &  \\
 \midrule
GPT 4 & N-Shot&  --& 0.72 \\
\citepalias{rajpoot-parikh-2023-gpt} &  N=5+4$\times{}$classes&  &  \\
 \midrule
PaLM 2$^2$ & 0-Shot$^2$&  --& 0.64 \\
\citepalias{aguda-etal-2024-large} &  1-Shot$^2$& -- & 0.67 \\
 &5-Shot$^2$  & -- & 0.69 \\
\midrule
\textbf{G-2.5-F,GPT-4o} &\textbf{Gen-Ref} & \textbf{0.51} & \textbf{0.61} \\
\textbf{GPT-o4,GPT-4.1} & \textbf{Hier-Multi} & \textbf{0.45} & \textbf{0.62}  \\
\textbf{G-2.5-F,GPT-4o} & \textbf{Dyn-Ex} & \textbf{0.50} & \textbf{0.64} \\
\bottomrule
\end{tabular}
\vspace{1mm}
\raggedright
\footnotesize{1) Pretrained specifically for relation extraction.\\
2) Reported temperature was 0.7; our methods consistently used temperature = 0. Also \citet{aguda-etal-2024-large} used few-shot prompts with examples tailored to the specific entity-pair types. In contrast, we used randomly selected examples.}
\end{table}

Table~\ref{tab:comparison_prior_work3} reports relation classification results on the SemEval dataset. Our Generator-Reflection architecture achieves a strong F1 Micro of 0.76, closely approaching the performance of fine-tuned models such as T5-large (0.80) and Mistral-7B (0.77), despite not relying on any task-specific fine-tuning. Compared to prompting-based baselines like GPT-4 and Llama-2, our method delivers a substantial performance boost, highlighting the advantages of structured multi-agent collaboration over single-shot prompting. While R-BERT reports the highest F1 Macro at 0.89, it benefits from pretraining explicitly tailored for relation classification, whereas our approach uses general-purpose models and maintains competitive performance without task-specific optimization, underscoring its adaptability and effectiveness in zero- or low-supervision settings.

\begin{table}[t]
\centering
\small
\caption{Relation classification performance on the SemEval dataset. Bold rows are our results. \\
G-2.5-P = Gemini 2.5 Pro; Gen-Ref = Generator-Reflection architecture.}
\label{tab:comparison_prior_work3}
\begin{tabular}{llcc}
\toprule
\textbf{Model} & \textbf{Setting} & \textbf{F1-Mac} & \textbf{F1-Mic} \\
\midrule
T5-large & Fine-tuning & -- & 0.80 \\
Mistral-7B & Fine-tuning & -- & 0.77 \\
\citepalias{Efeoglu2024RelationEW} &  &  &  \\
 \midrule
GPT-4 & N-Shot&  --& 0.65 \\
GPT-3.5 & N-Shot&  --& 0.61 \\
Llama-2 & N-Shot&  --& 0.57 \\
\citepalias{Guo2025BridgingGA} & N=1$\times$classes &  & \\
 \midrule
R-BERT$^1$ & Pre-trained$^1$ &  0.89& -- \\
\citepalias{Wu2019EnrichingPL} & BERT&  & \\
\midrule

\textbf{G-2.5-P,GPT-o3} &\textbf{Gen-Ref} & \textbf{0.52} & \textbf{0.61} \\
\textbf{G-2.5-F,GPT-4o} & \textbf{Hier-Multi} & \textbf{0.55} & \textbf{0.61} \\
\textbf{G-2.5-F,GPT-4o} & \textbf{Dyn-Ex} & \textbf{0.59} & \textbf{0.64} \\
\bottomrule
\end{tabular}
\vspace{1mm}
\raggedright
\footnotesize{1) Pretrained model
specifically optimized for relation classification.}
\end{table}

Table~\ref{tab:results_summary} compares our internal configurations across the CORE, REFinD, and SemEval datasets. On CORE, prompting with Gemini 2.5 Flash shows modest gains from zero- to few-shot, with performance plateauing at 5-shot. Switching to GPT-4o provides stronger zero-shot results but offers less improvement with additional shots. In contrast, our multi-agent setups, including Generator-Reflection, Hierarchical Multi-Agent, and Dynamic-Example agents, consistently outperform simple prompting approaches. Notably, the Generator-Reflection architecture with Gemini 2.5 Pro and GPT-o3 achieves the highest F1 Micro of 0.76 on CORE. On REFinD, the Dynamic-Example agent yields the best F1 Micro (0.64), showcasing its ability to enhance performance through more contextually adaptive example selection. For SemEval, this architecture again leads, achieving top scores in both macro and micro F1. These results underscore the generalizability and robustness of our multi-agent strategies, which deliver strong performance across datasets without relying on fine-tuning or heavily engineered prompts.

\begin{table*}[h]
\centering
\small
\caption{Our results across three datasets under different model and architecture configurations. Bold numbers are the highest scores for each dataset.\\
Gen = Generator agent; Ref = Reflection agent; Orc = Orchestrator agent; Spe = Specialist agent; Hier-Multi = Hierarchical Multi-Agent architecture; Cls: Classification agent; Ex: Example Selection agent; Dyn-Ex = Dynamic-Example Generator agent; G-2.5-F = Gemini 2.5 Flash; G-2.5-P = Gemini 2.5 Pro; GPT-o3 = GPT o3 mini; Sonn-4 = Claude Sonnet 4; GPT-o4 = GPT o4 mini; GPT-4.1 = GPT 4.1 mini.}
\label{tab:results_summary}
\begin{tabular}{llcccccc}
\toprule
\textbf{Models} & \textbf{Architecture} & \multicolumn{2}{c}{\textbf{CORE}} & \multicolumn{2}{c}{\textbf{REFinD}} & \multicolumn{2}{c}{\textbf{SemEval}} \\
\cmidrule(lr){3-4} \cmidrule(lr){5-6} \cmidrule(lr){7-8}

 &  & F1 Macro & F1 Micro & F1 Macro & F1 Micro & F1 Macro & F1 Micro \\
\midrule
 & Zero Shot & 0.712& 0.703& 0.457& 0.553& 0.466& 0.566\\
    G-2.5-F             & 3 Shot    & 0.730& 0.741& 0.467& 0.562& 0.477& 0.580\\
                 & 5 Shot    & 0.721& 0.724& 0.493& 0.602& 0.467& 0.576\\
\midrule
    & Zero Shot & 0.749& 0.716& 0.498& 0.622& 0.406& 0.493\\
     GPT-4o            & 3 Shot    & 0.757& 0.718& 0.475& 0.575& 0.394& 0.474\\
                 & 5 Shot    & 0.747& 0.703& 0.489& 0.591& 0.429& 0.537\\
\midrule
Gen: G-2.5-F , Ref: GPT-4o &  & 0.753& 0.735& \textbf{0.507}& 0.607& 0.464& 0.573\\
Gen: G-2.5-P , Ref: GPT-o3 & Gen-Ref & 0.767& \textbf{0.759}& 0.481& 0.586& 0.516& 0.606\\
Gen: Sonn-4 , Ref: GPT-o3 &  & \textbf{0.771}& 0.744& -- & -- & -- & --\\
\midrule
Orc: G-2.5-F, Spe: GPT-4o & Hier-Multi & 0.738& 0.733& 0.405& 0.550& 0.549& 0.611\\
Orc: GPT-o4, Spe: GPT-4.1 &  & 0.697& 0.749& 0.453 & 0.620 &  0.478 & 0.580 \\
\midrule
Gen\&Cls:G-2.5-F, Ex:GPT-4o & Dyn-Ex & 0.752& 0.720& 0.498& \textbf{0.642}& \textbf{0.591}& \textbf{0.635}\\
\bottomrule
\end{tabular}
\end{table*}

Table~\ref{tab:orchestrator_performance} presents the Orchestrator Agent’s routing performance within the Hierarchical Multi-Agent architecture, measured by F1 scores across Specialist Agents on three datasets. On CORE, both model configurations demonstrate consistently high routing accuracy across all specialists, indicating that the Orchestrator effectively distinguishes among domain-specific relation types. For REFinD, the first configuration achieves particularly strong routing performance, with scores up to 0.98, reflecting the dataset’s clear domain structure. In the second configuration, routing accuracy is more varied, suggesting increased sensitivity to model choice in complex financial contexts. On SemEval, routing F1 scores are slightly lower across all specialists, suggesting a more diffuse or overlapping distribution of relation types in this general-domain benchmark. Nonetheless, the Orchestrator maintains solid performance, highlighting its robustness even in more ambiguous classification scenarios.

\begin{table*}[h]
\centering
\small
\caption{F1 scores of the Orchestrator Agent in the Hierarchical Multi-Agent architecture, showing routing accuracy to each Specialist Agent across datasets.\\
Orc = Orchestrator; Spe = Specialist; G-2.5-F = Gemini 2.5 Flash; GPT-o4 = GPT o4 mini; GPT-4.1 = GPT 4.1-mini.}
\label{tab:orchestrator_performance}
\begin{tabular}{lccc ccc ccc}
\toprule
\textbf{Model Configuration} 
& \multicolumn{3}{c}{\textbf{CORE}} 
& \multicolumn{3}{c}{\textbf{REFinD}} 
& \multicolumn{3}{c}{\textbf{SemEval}} \\
\cmidrule(lr){2-4} \cmidrule(lr){5-7} \cmidrule(lr){8-10}
& Spe1 & Spe2 & Spe3
& Spe1 & Spe2 & Spe3
& Spe1 & Spe2 & Spe3 \\
\midrule
Orc: G-2.5-F, Spe: GPT-4o & 0.88 & 0.84 & 0.86 & 0.94 & 0.98 & 0.91 & 0.82 & 0.79 & 0.74 \\
Orc: GPT-o4, Spe: GPT-4.1 & 0.88 & 0.84&0.87 & 0.71& 0.60&0.88 & 0.84& 0.79& 0.74\\
\bottomrule
\end{tabular}
\end{table*}


\section{Conclusion}
This work introduces a systematic evaluation of three multi-agent architectures designed to enhance relation classification using LLMs. By comparing reflective critique (Generator-Reflection), hierarchical specialization (Hierarchical Multi-Agent), and adaptive example construction (Dynamic-Example Generator), we highlight how distinct modes of agentic coordination can complement LLM capabilities in structured tasks. Across three domains—financial (REFinD), scientific (CORE), and general-domain (SemEval)—our approaches outperform prompting baselines and demonstrate competitive results with fine-tuned systems, despite requiring no task-specific training. The modularity of these architectures enables flexibility, transparency, and generalization, particularly in zero- and few-shot scenarios. Future work will extend this framework to more complex tasks, such as document-level relation extraction, cross-lingual adaptation, and integrating multi-modal signals.



\bibliography{custom}
\bibliographystyle{acl_natbib}

\end{document}